\documentclass[conference]{IEEEtran}
\IEEEoverridecommandlockouts
\usepackage{cite}
\usepackage{amsmath,amssymb,amsfonts}
\usepackage{algorithmic}
\usepackage{graphicx}
\usepackage{textcomp}
\usepackage{tabularx}
\usepackage[table,xcdraw]{xcolor}
\usepackage{xcolor}
\usepackage{amssymb} 
\usepackage{bbding}
\usepackage{mathrsfs}
\def\BibTeX{{\rm B\kern-.05em{\sc i\kern-.025em b}\kern-.08em
    T\kern-.1667em\lower.7ex\hbox{E}\kern-.125emX}}
\begin{document}

\title{CodingTeachLLM: Empowering LLM's Coding Ability via AST Prior Knowledge}

\author{\IEEEauthorblockN{1\textsuperscript{st} Zhangquan Chen}
\IEEEauthorblockA{\textit{Beijing OneXOne Tech Co., Ltd} \\
\textit{Tsinghua University}\\
Beijing, China \\
chenzhangquan@hibug.com}
\and
\IEEEauthorblockN{2\textsuperscript{st} Chunjiang Liu}
\IEEEauthorblockA{\textit{Beijing OneXOne Tech Co., Ltd} \\
Beijing, China \\
liuchunjiang@hibug.com}
\and
\IEEEauthorblockN{3\textsuperscript{st} Haobin Duan}
\IEEEauthorblockA{\textit{Beijing OneXOne Tech Co., Ltd} \\
Beijing, China \\
duanhaobin@hibug.com}
}

\maketitle

\begin{abstract}
In this paper, we introduce CodingTeachLLM, a large language model (LLM) designed for coding teaching. Specially, we aim to enhance the coding ability of LLM and lead it to better teaching mode in education context. Thus, we propose an end-to-end prior-based three-phases supervised fine-tuned model, which is proved more competitive than traditional fine-tuning method. More specifically, our model realizes the structural disassembly and incremental guided output of educational knowledge. To this end, we robustify data classification of three types via a sampler and overlap estimation neural network, and inject the preprocessing datasets into pre-trained model in three batches for LORA fine-tuning. Then, we design a prior module couples system prompt, vector databases, and abstract syntax tree task segmentation. Finally, the compression method and regularization constraint are applied to the prior-based fine-tuned model, followed by text filter at the output end to obtain incremental guided results. Our model represents the first research effort to truly embody the tutor role with the features of abundant educational knowledge, step-by-step incremental guided outputs and non-disclosure of answers. Extensive experiments report that our model also achieves \textbf{state-of-the-art} in code abilities compared to open-source models, reaching an impressive \textbf{75.10\%} on the HumanEval (@pass 1) benchmark. Additionally, our model maintains strong conversational capabilities, with the 13B quantized version achieving scores of \textbf{56.34}, \textbf{50.60}, and \textbf{45.27} respectively on the MMLU, C-Eval, and AGIEval (5 shot) dialogue evaluation benchmarks.
\end{abstract}

\begin{IEEEkeywords}
Three-phases Supervised Fine-tuned Model, Prior Module, Incremental Guided Output
\end{IEEEkeywords}

\section{Introduction}
Mathematician Markov proposed the Markov chain in 1906, information theory founder Shannon proposed the concept of information entropy in 1948, and linguist Chomsky put forward the theory of transformational-generative grammar in 1957, all of which had significant impacts on the formation of the language models. Traditional language models adopt statistical approaches to describe the probability of a sequence of characters forming sentences, such as the widely-used N-gram model proposed in 1980. In 2003, Bengio et al. \cite{bengio2000neural} introduced the first neural language model named Feedforward Neural Network(FFNN), opening a new era in language modeling. Although the performance of FFNN was superior to N-gram models, training was still expensive and inefficient. In 2010, Mikolov et al. \cite{mikolov2010recurrent} proposed the Recurrent Neural Network Language Model (RNN), which significantly outperformed FFNN in terms of perplexity. But RNN suffered from the vanishing or exploding gradient problem, leading to slow training or unbounded parameter values. In 2012, Sundermeyer et al. \cite{sundermeyer2012lstm} further proposed the Long Short-term Memory Recurrent Neural Network Language Model (LSTM-RNN), which addressed the gradient problem of the recurrent neural network. Although the performance of LSTM was promising, training models on large-scale corpora were time-consuming. In 2017, the Google Brain team abandoned the recurrent neural network structure and introduced the Transformer\cite{vaswani2017attention}, a self-attention-based neural network architecture. Transformer architecture really revolutionized the field, allowing for greater parallelism, faster training, and versatility. In 2018, OpenAI introduced the GPT (Generative Pre-Training) language model with 117 million parameters\cite{radford2018improving} , which used part of the decoder of the multi-layer Transformer (unidirectional self-attention mechanism) as the language model and refreshed 9 records in 12 Natural Language Processing(NLP) tasks. Soon after, Google released the BERT language model\cite{devlin2018bert} via the encoder of Transformer architecture. BERT used a bidirectional self-attention mechanism instead, which refreshed 11 records in 12 NLP tasks. Consequently, many subsequent language models were developed based on the open-source BERT, and the Transformer architecture began to dominate the NLP field. The introduction of GPT and BERT models signaled the advent of the era of large-scale AI models and the transition of language models to large-scale language models. More specifically, large language models(LLMs) refer to models with numerous parameters, extensive training data, and intended for natural language processing tasks. 

Though it is a piece of cake for universal LLMs to solve general natural language problems, they still face various challenges in domain-specific context. This is mainly because of the limitation of professional knowledge and the weak cognition in specialized domains. Professional model as a tutor in the education context is a huge challenge. As a qualified tutor, a large number of expertise is needed, and tutor is expected to break down all of the knowledge points to output logically. Only in this way can the students(users) be more receptive. For general LLMs, they can output complete and detailed text via the overall text probability but they cannot decompose the knowledge logic based on teachers' cognition. Neither could they dress as a truly tutor to guide students in independent learning. Based on large-scale pre-trained models such as GPT, supervised fine-tuning(SFT) can easily be conducted to adapt to downstream tasks including education domain. Hence, this paper proposes an end-to-end SFT educational model via an a priori module, with the main contributions as follows:
\begin{enumerate}
\item Introduce a neural network structure for LLM-dataset preprocessing, which achieves high-quality dataset via random sampling and overlap estimation network.
\item Demonstrate the superiority of the three-phases LORA fine-tunin. Compared with the traditional SFT method, stepwise dataset injections for fine-tuning can significantly improve the model's performance in specific domains.
\item Design a composite prior module, integrating vector databases, abstract syntax tree(AST), and efficient system prompt to implement strong correlation constraints associated with the tutor role.
\item Optimize the educational model through regularization constraints, model compression and pruning and text filter, proving the feasibility solutions in the education context.
\item Truly embody the essence of a tutor, and firmly achieve SOTA in coding capabilities among all of the open-source LLMs. Also demonstrate the extraordinary accuracy and robustness in multiple comparative experiments.
\end{enumerate}
\section{Related Works}
\subsection{Generative Large Language Model}

In recent years, the field of generative artificial intelligence has made numerous breakthroughs. LLMs can not only satisfy the need for individuals to quickly obtain answers but also provide personalized learning and support for human-computer interaction. Taking GPT models as an example, before the first-generation GPT-1 model\cite{radford2018improving} was proposed, most deep learning methods tended to adopt supervised training and learning approaches. These required a large amount of manually annotated high-quality data. However, the massive data annotation costs greatly limited the performance ceiling and generality of these methods across various tasks, leading to the emergence of two new paradigms: unsupervised and supervised fine-tuning. The GPT-2 model\cite{radford2019language} placed a greater emphasis on the model's generality and multi-task learning, adapting to multiple different downstream tasks without changing the model's structure by introducing additional input information. The GPT-3 model\cite{brown2020language} further expanded the model. Although GPT-3 demonstrated powerful few-shot learning capabilities, its method of completing tasks still required several task example data. In early 2022, the InstructGPT model\cite{ouyang2022training} followed a similar technical approach to the current chatGPT by utilizing supervised fine-tuning, reward model training, and proximal policy gradient algorithms. Subsequently, Copilot achieved vertical task implementation in the programming domain. Focusing on open-source models, GPT spurred the development of models like Alpaca\cite{taori2023alpaca}, which used 52K data distilled from the OpenAI API in the ChatRWKV project\cite{radford2019language}. Alpaca explored non-mainstream Transformer model architectures like the RNN and achieved complete pre-training and alignment of human preferences through fine-tuning. The Vicuna\cite{chiang2023vicuna} model trained LLaMA\cite{touvron2023llama} on human and ChatGPT conversation data from ShareGPT, achieving performance close to Google Bard's. Koala\cite{geng2023koala} used distilled data and open-source dialogue data for fine-tuning LLaMA, obtaining results close to ChatGPT.

Furthermore, bilingual models like ChatGLM\cite{zeng2022glm} and Baichuan\cite{yang2023baichuan} achieved breakthroughs in Chinese-English LLMs. In the process of LLM development, the open-source community introduced technologies like Self-Instruction, Context Distillation, HIR, the Chain of Thoughts (COT)\cite{wei2022chain}, instructional fine-tuning\cite{wang2022self}, and distributed training frameworks, laying the foundation for the development of LLMs.
\subsection{Specialized Vertical Model}
Although LLMs demonstrate impressive capabilities in general domains, their lack of subject-matter expertise becomes apparent when applied to specialized vertical domains. For instance, we can find dedicated language models suitable for various areas, such as ChatDoctor\cite{li2023chatdoctor} for the healthcare domain, Chat-Law\cite{cui2023chatlaw} for the legal domain, and FinGPT\cite{yang2023fingpt} for the finance domain. These niche domains inherently require models to possess comprehensive knowledge in the field to handle relevant queries, especially when assisting actual users in real-life situations.

The education domain presents an even more challenging scenario, as a professional tutor needs to have pedagogical knowledge and guidance abilities and present specialized knowledge in a way that is accessible to students. Baladn et al.\cite{baladn2023retuyt} adapted open-source LLMs to generate better teacher responses in the BEA 2023 shared task\cite{tack2023bea}. Educhat\cite{dan2023educhat} is pre-trained on a diverse education corpus to ensure the alignment with educational abilities. IBL-Tutoring is a tutor model that is fine-tuned against Mistrallite to respond like a professional teacher. But challenges still exist, these kinds of model only realize the role of teacher through the fine-tuning of a single educational corpus. But they cannot produce instructive output, let alone output in a way that is easy for students to understand. The essence of this is the overlaying of educational expertise on a generic LLM.

\section{Methodology}
Although LLMs have been explored in general and a subset of vertical fields, the outputs of the models is full throughput and cannot provide step-by-step incremental guided answers to user's questions. For example, when we firstly put a role system prompt to chatGPT "I am a student, then please play the role of a tutor and guide me step by step to implement the algorithm, don't tell me the answer directly!". And then we issue the question "Implement a bubble sort code using python", chatGPT will directly and accurately output all the answers "of course! Bubble sorting is a simple but less efficient sorting algorithm that performs sorting by comparing adjacent elements and gradually switching positions. Here is an example code that implements bubble sort:\textbackslash n def bubble\_sort(arr):\textbackslash n n = len(arr)\textbackslash n \textbackslash n for i in range(n):\textbackslash n \# In each iteration, perform n-i-1 comparisons\textbackslash n for j in range(0, n-i-1):\textbackslash n \# If the current element is greater than the next element, swap their positions\textbackslash n if arr{[}j{]} \textgreater{} arr{[}j+1{]}:\textbackslash n arr{[}j{]}, arr{[}j+1{]} = arr{[}j+1{]}, arr{[}j{]}\textbackslash n \textbackslash n return arr". But what we expect is that the LLM can truly understand the role of a tutor, guiding students step by step to solve the problem themselves without giving the standard answer. It would be better if the model can divide the total task of bubble sort implementation into some specific knowledge like loop $for$, function definition $def$, conditional statement $if$, etc., and gradually guide students to realize the corresponding code with independent thinking ability. GPT-4 with up to 1750B parameters can achieve a certain degree of guidance according to appropriate instructions. Is is mainly due to the powerful data volume and large model parameters behind GPT-4. Such a huge model inevitably leads to the low inference efficiency and the enormous requirement of GPU resources. In this case, how to adapt the vertical task of step-by-step incremental guided output for small and medium-sized models has become an challenging problem. 
\begin{figure*}[htbp]
	\centering
	\includegraphics[width=\linewidth,scale=1.00]{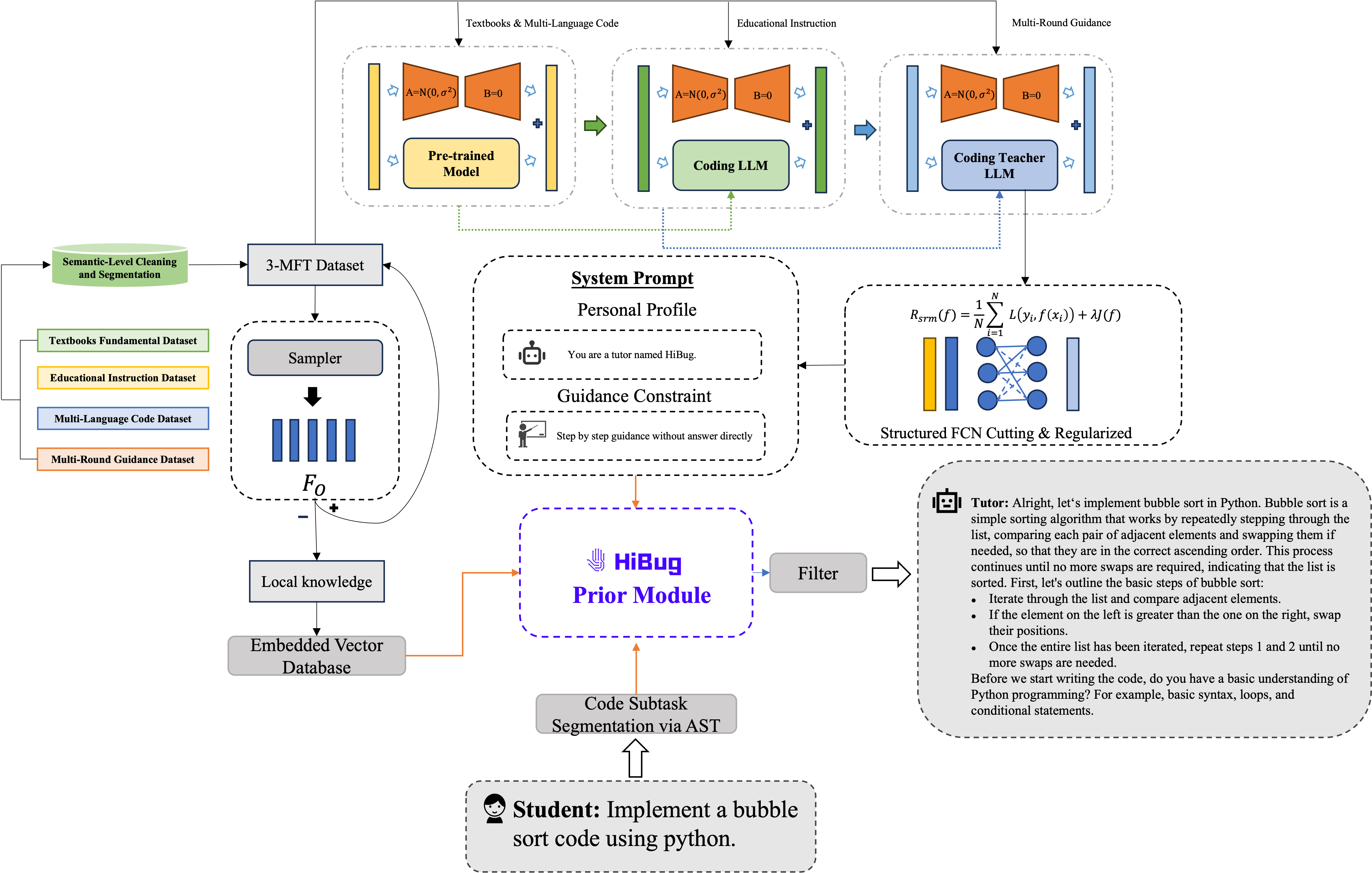}
	\caption{Overview of our method. \textbf{End-to-End: A supervised learning network coupled peripheral algorithm modules}: data segmentation and distillation algorithm via powerful LLM, overlap estimation neural network used for data pre-processing, three phases fine-tuning of seq2seq language model, supervised model fine-tuning under efficient regularization constraints, prior module combining AST and vector database, better designed filter for dynamic noise.}
	\label{FigureOne}
\end{figure*}

Thus, we propose a incremental guided outputs system with strong robustness, high accuracy, quick inference and low GPU resource consumption. On the pre-trained model of Transformer architecture, data distillation is intrudued based on specially processed corpus and overlap estimation network. Firstly, the first round of basic ability training of the model is carried out. Then, the second round of teacher corpus role alignment is realized. And the third alignment is performed on the guidance corpus. At the same time, a prior component is constructed by integrating local knowledge database, code logic tree and system prompt module. The lightweight storage of multi-round conversations is realized by vector database as well as well. The self-feedback filter is carried out to suppress the output noise of the model, so as to realize the step-by-step incremental guided outputs question answering system.

\subsection{Dataset}\label{AA}
In our research, we gather a vast amount of data divided into four parts. \textbf{Textbooks Fundamental Dataset:} in order to achieve a more natural human-computer interaction, we collect a large volume of bilingual instruct tuning data from reputable open-source repositories like BELLE\cite{ji2023exploring}, GPT4All, FLANCoT and  Alpaca. These general-purpose datasets contain a large number of phrases, sentences, and paragraphs and are suitable for tasks such as question answering, classification, translation, text summary generation, and reading comprehension. Besides, we also add some cleaning datasets ranging from encyclopedia, forum, web corpus, conversation text and so on collected ourselves, so as to develop the general ability of LLM in both Chinese and English. \textbf{Educational Instruction Dataset:} this part contains Chinese/America middle and high school exam textbooks and online question bank data, and we also use 180,000 pieces of Chinese/English cultural knowledge to further enrich our model. At the same time, we also added the corpus of tone and dialogue mode of different kinds of teacher roles, so as to enhance the cognition of the model in the field of education. \textbf{Multi-Language Code Dataset:} code datasets can improve the expertise of the model and promote the logic ablility of model as well. Our code datasets not only comes code accumulation within the company, but also code open-source communities like github, huggingface, etc. Later, we use GPT-4 for raw data distillation and expansion, producing high-quality code data based on some segmentation and extraction algorithms.
\textbf{Multi-Round Guidance Dataset:} We source high-quality multi-turn dialogue data from ShareGPT10, MOSS\cite{sun2023moss}, BELLE\cite{ji2023exploring}, LIMA\cite{zhou2023lima}, and COIG\cite{zhang2023chinese}. For our own one-round question-and-answer dataset, we split it into multi-round conversations. A general task is divided into many subtasks, and each round of dialogue assistant's answer illustrates only one of the substeps.
\subsection{Overlap Estimation Network}
The LLM provides excellent output when there is a strong correlation between the local knowledge base and the finetune dataset. Therefore, the local knowledge base data should ideally be tightly coupled to the finetune data. We randomly sampled all the data (Multiphase Fine-Tuning dataset, also named MFT-dataset) $M = \left \{  X_{1},X_{2},X_{3}...,X_{n} \right \} $ to obtain discrete data samples $S = \left \{  X_{1},X_{2},X_{3}...,X_{m} | m\le \frac{n}{2} \right \} $. For the data sample $S$, we estimate its correlation with the data $M-S$. If the correlation is greater than the threshold $T$, we enter the local knowledge base. Otherwise, the data would return the MFT-dataset. So the final local knowledge dataset $S^{'} = \left \{  X_{1},X_{2},X_{3}...,X_{k} | k\le n,F_o(M-S^{'},S^{'})>T \right \} $, and MFT-dataset $M^{'} = M - S^{'}$. Overlap network $F_o$ is a deep regression network structure, consisting of 3 convolutional layers and 2 fully connected networks. It is obtained by small sample training for similar data sets.

\subsection{Three-phases LORA Fine-tuning}
We used model LLAMA2-34B\cite{touvron2023llama2} as our pre-trained model and used LORA\cite{hu2021lora} method for three-stages fine-tuning. Through a large number of experiments, we found that the result of single step fine-tuning is terribly bad with very strong hallucination, forgetting and overfitting effects. However, fine-tuning code data, educational awareness, and guidance in three-stages procedure has a significantly improve model capability. The tutor role is well realized with the model code ability is improved as well. At each phase, we froze the original model parameters and superimposed low-rank decomposition weights to fine-tune vertical tasks. Specifically, for a pre-trained weight matrix $W_0 \in \mathbb{R}^{d \times k}$, we constrain its update by representing the latter with a low-rank decomposition $W_0+\Delta W=W_0+B A$, where $B \in \mathbb{R}^{d \times r}, A \in \mathbb{R}^{r \times k}$, and the rank $r \ll \min (d, k)$. During training, $W_0$ is frozen and does not receive gradient updates, while $A$ and $B$ contain trainable parameters. For $h=W_0 x$, our modified forward pass yields:
\begin{equation}
h=W_0 x+\Delta W x=W_0 x+B A x
\end{equation}

We illustrate our reparametrization in \ref{FigureOne}. We use a random Gaussian initialization for $A$ and zero for $B$, so $\Delta W=B A$ is zero at the beginning of training. We then scale $\Delta W x$ by $\frac{\alpha}{r}$, where $\alpha$ is a constant in $r$. When optimizing with Adam, tuning $\alpha$ is roughly the same as tuning the learning rate if we scale the initialization appropriately. As a result, we simply set $\alpha$ to the first $r$ we try and do not tune it.

The data used at each phase and the four classifications of the original raw datasets are surjective. \textbf{First phase:} use textbooks data and code data to fine-tune the LLAMA2-34B with original weights. The former operation is to deepen the bilingual ability of English and Chinese, and the latter operation is to make the model more focused on the code field. Generate $LLM^{1}$. \textbf{Second phase:} $LLM^{1}$ is fine-tuned with the education data to improve the educational cognition of the model. Generate $LLM^{2}$. \textbf{Third phase:} $LLM^{2}$ is fine-tuned with the guided multi-round dialogue data to enhance the step-by-step incremental guided ability. Generate $LLM^{3}$. In this way, $LLM^{3}$ can intelligently stop output at the appropriate position as a tutor, carry out multiple rounds of dialogue multiple throughput, and avoid a single full throughput with the outflow of answers.

\subsection{Structured FCN Cutting and Regularized}
One thing to emphasize is that we use a special network structure between $LLM^{2}$ and $LLM^{3}$. In order to make the model more guided, we firstly use the regular term in structured risk function for constraints:
\begin{equation}
R_{s r m}(f)=\frac{1}{N} \sum_{i=1}^N L\left(y_i, f\left(x_i\right)\right)+\lambda J(f)
\end{equation}

Where $J(f)$ represents the complexity of the model and is a functional defined on the hypothesis space $\mathcal{F}$. The more the model throughput, the worse the guidance capability, and the larger the $J(f)$ will be. Structured pruning in the finetune process mainly focuses on batchnorm. If the scaling factor in the batchnorm layer behind a channel is small enough, then the channel is of low importance and the dashed line is pruned:
\begin{equation}
\hat{z}=\frac{z_{\text {in }}-\mu_\beta}{\sqrt{\delta_\beta^2+\epsilon}} ; z_{\text {out }}=\gamma \hat{z}+\beta
\end{equation}

$\gamma$ is the channel scaling factors. Therefore, the network is more sparse, the number of parameters is reduced, and the interference of non-tutor neurons can be effectively avoided.

\subsection{Prior Module}
The prior information module is the pivot in the whole system, which combines the trained model and a series of pre-processing algorithms. Bayes criterion can be used to strengthen transformer inference logic. More deterministic prior distributions are more beneficial to posterior inference, thus we design this module to enhance the ability of later inference. The prior module includes pre-system prompt, vector database and subtask segmentation. \textbf{Pre-system prompt:} the system prompt provides custom roles for LLMs. We mainly give two parts, personal profile and guide constraint, to increase the recognition of tutor roles. \textbf{Vector database:} we use text2vec-base-chinese to implement embedding, building a dense vector efficient similarity retrieval and clustering engine based on Faiss. Therefore, the local knowledge base is converted to vector database to realize lightweight storage and efficient similarity search. \textbf{Subtask segmentation:} an abstract syntax tree is a tree-like data structure which represents the syntax structure of code through nodes and edges. Based on AST, we can obtain the semantics of code, and then carry out code segmentation based on semantics and structure. Thus, the complete code task is divided into many sub-task modules to assist the implementation of the guidance function.
\subsection{Inference Procedure}
When the user enters prompt "Implement a bubbling sort code in python". First of all, the system will be divided into many different detailed code tasks through the AST for this task: including storage arrays, judgment statements, loop statements, etc. At the same time, the correlation between prompt and vector database information is calculated based on cosine similarity, and the part with high correlation is regarded as a prior knowledge. Finally, original prompt with logical structure statements and prior knowledge information will be carried into the trained model. So far, the trained model with the appropriate system prompt preset will begin the inference of the tutor mode. It should be acknowledged that the output of direct inference contains some noise and unrobust factors. Moreover, it is found through experiments that with the increase of context length, the model will weaken the system prompt and focus on the current information. Therefore, if you constantly ask the model when you have a lot of context, the model may reveal the answer with bubble sort code we have mentioned above, thus lose the meaning of the tutor role. To solve this problem, we design a filter to drop out the noise that affects the output. In simple terms, if this part of the noise is negatively related to the model mentor role, then we will suppress this part of the noise. And, following the Markov chain principle, if the output always has full throughput, we will revert to the previous derivation. Compared with the full throughput of the traditional LLM, our system can induce the output of multiple rounds of dialogue in batches, realizing the true tutor role in a certain sense.

\section{Experimental Results}
In this section, we will evaluate the superiority of our model over other models on various datasets, demonstrating excellence in both code-related tasks and dialogue tasks. Additionally, when necessary, the model can switch to the role of a guided tutor.
\subsection{Data Preprocessing Via Overlap Estimation Network}
To quantify the robustness and reliability of our overlap estimation network, we conducted an analysis using a subset of preprocessed data. Specifically, we randomly sampled 10 MFT datasets and 10 Local Knowledge datasets, which are denoted as follows: $M_1, M_2, M_3 ... M_10$ and $L_1, L_2, L_3 ... L_10$.

\begin{figure*}[htbp]
	\centering
	\includegraphics[width=\linewidth,scale=1.00]{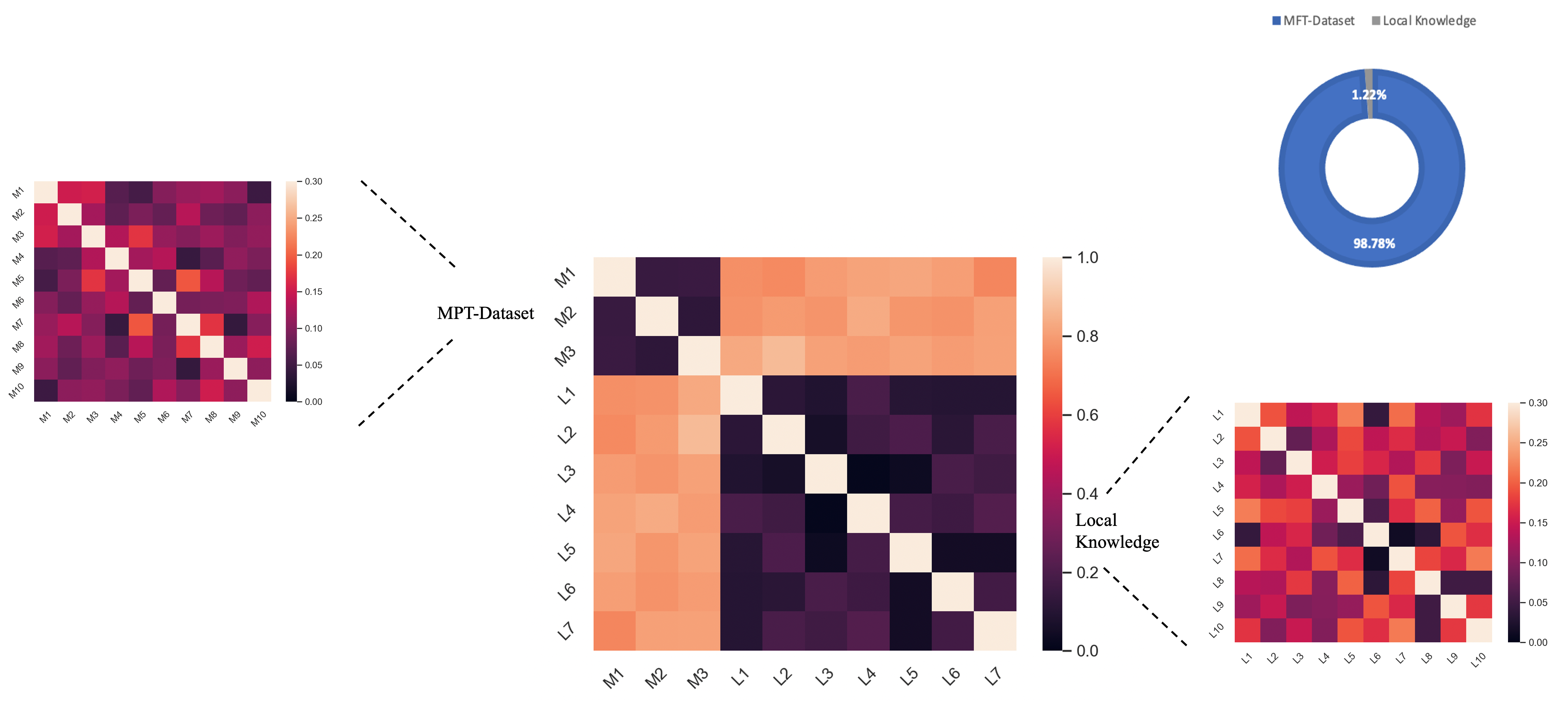}
	\caption{Heat map of oure data preprocessing results via overlap network. 10 samples were randomly selected, where M represents MFT-dataset and L represents Local Knowledge.}
	\label{figure_map}
\end{figure*}

For the sampled data, we encoded the data into vector space for subsequent calculation of cosine similarity. As shown in the figure \ref{figure_map}, we firstly present the cosine vector similarity between each pair of 10 samples within the MFT dataset and the Local Knowledge dataset, taking $M_1$ and $M_2$ as the example(n represents the dimension of the vector $M_k$,):

\begin{small} 
\begin{equation}
\rho =\cos (\theta)=\frac{M_1 \cdot M_2}{\|M_1\|\|M_2\|}=\frac{\sum_{i=1}^n M_{1i} \times M_{2i}}{\sqrt{\sum_{i=1}^n\left(M_{1i}\right)^2} \times \sqrt{\sum_{i=1}^n\left(M_{2i}\right)^2}}
\end{equation}
\end{small} 

Figure \ref{figure_map} shows the result using heat map. In the complete dataset, the MFT dataset accounts for approximately \textbf{98.78\%}, while the Local Knowledge dataset accounts for approximately \textbf{1.22\%}. The cosine similarity between each pair of samples within the dataset is below \textbf{0.3}, indicating that the data used for fine-tuning is of high quality and highly independent from each other. The same applies to the internal data of the Local Knowledge dataset. However, when comparing the MFT dataset and the Local Knowledge dataset, such as $M_1, M_2$ and $L_1, L_2...L_8$, the cosine similarity between these types of data exceeds \textbf{0.8}, indicating a high degree of correlation. This is advantageous for incorporating the Local Knowledge dataset as a prior module in the model structure.

\subsection{Fine-tune Training And Compression}
As mentioned in the previous section, we fine-tune our model based on LLAMA2-34B for three rounds on a well-processed training set, including general domain knowledge, code, education, and multi-turn dialogue data. Meanwhile, the powerful prior self-supervised processing model strengthens the model's learning and self-feedback capabilities.

The model we trained consists of two main parts: \textbf{ours-34B} is the part of full parameter training, and \textbf{ours-13B} is the result obtained after quantization, compression and pruning of the model.
\subsection{Coding Ability}
\textbf{HumanEval:} HumanEval\cite{chen2021evaluating} is a dataset designed for evaluating the performance of code generation models, introduced by OpenAI in 2021. This dataset contains 164 handcrafted programming problems, each of which includes a function signature, a documentation string (docstring), a function body, and several unit tests. These problems cover various aspects such as language understanding, reasoning, algorithms, and simple mathematics. The generated code is considered correct only if it passes all the relevant unit tests.

\ref{table:tab_code} reports the results of coding ability on the benchmark HumanEval. In terms of single-pass success rate, our model's accuracy is \textbf{0.94\%} higher than the open-source SOTA CodeFuse-CodeLlama-34B, and outperforms GPT-4 (zero-shot) by \textbf{12.09\%}.
\begin{table}[t!]
\caption{Evaluate The Coding Ability Of Different Model On HumanEval Dataset. The \textbf{Best} Is Highlighted.}\label{table:tab_code}
\centering
\begin{tabular}{lc}
\hline
\textbf{Model} & \textbf{HumanEval(pass@1)} \\ \hline
StarCoder-15B\cite{li2023starcoder}             & 33.60\%                    \\
OctoCoder\cite{muennighoff2023octopack}                & 46.20\%                    \\
GPT-3.5(zero-shot)      & 48.10\%                    \\
CodeLlama-34B\cite{roziere2023code}             & 48.80\%                    \\
PanGu-Coder2-15B\cite{christopoulou2022pangu}          & 61.60\%                    \\
GPT-4(zero-shot)         & 67.00\%                    \\
WizardCoder-Python-34B\cite{luo2023wizardcoder}   & 73.20\%                    \\
CodeFuse-CodeLlama-34B\cite{liu2023mftcoder}   & 74.40\%                    \\
\rowcolor[HTML]{E7E6E6} 
\textbf{ours-34B}                      & \textbf{75.10\%}           \\ \hline
\end{tabular}
\end{table}

\subsection{Chat Ability}
\textbf{MMLU\cite{hendrycks2020measuring}:} MMLU aims to measure the knowledge acquired during pre-training by specifically evaluating models in zero-shot and few-shot settings. This makes the benchmark more challenging and more akin to evaluating humans. The benchmark covers 57 subjects, including STEM, humanities, and social sciences. Its difficulty ranges from elementary to advanced professional levels, testing world knowledge and problem-solving abilities. The granularity and breadth of the subjects make the benchmark test an ideal choice for identifying model blind spots.

\textbf{C-Eval\cite{huang2023c}:} This is a Chinese-Enlish bilingual evaluation dataset, produced in collaboration with Shanghai Jiao Tong University, Tsinghua University, and the University of Edinburgh. It contains 13,948 multiple-choice questions, covering 52 different subjects and four difficulty levels.

\textbf{AGIEval\cite{zhong2023agieval}:} This benchmark selects 20 official, public, and high-standard qualification exams aimed at general human candidates, including general college entrance exams (such as China's Gaokao and the US SAT), judicial examinations, math competitions, and so on.

These types of datasets are mainly used to evaluate the bilingual conversation abilities of different model in both Chinese and English languages. To ensure fairness, the evaluation is conducted using the widely adopted 5-shot setting. \ref{table:tab_chat} reports the results of chat ability on the three benchmarks. Our model's conversational ability has achieved a SOTA position among the open-source models. It surpasses LLAMA2-13B by \textbf{2.27\%} on the MMLU evaluation benchmark, \textbf{41.34\%} on the C-Eval dataset, and \textbf{40.20\%} on AGIEval.
\begin{table}[]
\caption{Evaluate The Chat Ability Of Different Model On MMLU/C-Eval/AGIEval Dataset. The \textbf{Best} Is Highlighted.}\label{table:tab_chat}
\centering
\setlength{\tabcolsep}{1.2mm}{
\begin{tabular}{lccc}
\hline
\textbf{Model} & \textbf{MMLU(5-shot)} & \textbf{C-Eval(5-shot)} & \textbf{AGIEval(5-shot)} \\ \hline
C-Alpaca-13B      & 43.90 & 38.80 & 35.46 \\
ChatGLM2-6B\cite{zeng2022glm}     & 45.90 & 50.20 & 45.28 \\
Vicuna-13B\cite{chiang2023vicuna}   & 52.00 & 32.80 & 31.55 \\
LLaMA2-13B\cite{touvron2023llama2}   & 55.09 & 35.80 & 32.29 \\
\rowcolor[HTML]{E7E6E6} 
ours-13B                    & 56.34 & 50.60 & 45.27 \\
GPT-3.5 Turbo           & 68.54 & 51.10 & 46.13 \\
\textbf{GPT-4} & \textbf{83.93}        & \textbf{68.40}          & \textbf{63.27}           \\ \hline
\end{tabular}}
\end{table}

\subsection{Tutoring Ability}
Tutoring ability is a comprehensive application that integrates various capabilities such as guidance, long-context, and role-playing. For different models, we give the same instruction and the same system prompt: "You are a patient and cautious tutor. For the questions raised by users, you should guide and help them solve the problem themselves independently step-by-step without revealing the final result under any circumstances." To evaluate whether a model possesses educational tutoring ability, we judge it from three dimensions.Firstly, we consider the teacher-style(\textbf{Tutor Style}). We assess whether the model's output embodies the tone of a teacher, and observe whether it demonstrates a teacher's role cognition in multi-turn dialogues. Secondly, we evaluate the guidance provided in the output(\textbf{Guided Output}). Instead of directly telling users the result, we expect the model to gradually lead users towards the final result, achieving step-by-step batch processing rather than a single full-batch processing. Lastly, we evaluate the answer's disclosure(\textbf{w/o Answer}). As the conversation context grows, we observe whether the model has forgot its previous setting and inadvertently reveals the answer, such as directly outputting code or calculation results, which would be detrimental to nurturing independent thinking abilities among students. 

\ref{table:tab_tutor} reports the results of three different categories of models in tutoring tasks. The top category consists of open-source generative code-based LLM, the middle one represents open-source mainstream Chinese-English bilingual chat LLM, and the bottom category comprises closed-source commercial LLMs. We can see that code-based LLM barely possess role-playing capabilities, losing the majority of their chat abilities. The open-source chat LLM have some role-playing talents and can act as a tutor to a certain extent, but they fail to genuinely understand guidance, let alone provide it in multi-turn dialogues. Commercial LLMs (mostly around 175B parameters, with GPT-4 reaching 1750B) can convincingly play the role of a teacher. However, only GPT-4 and Claude-2 can comprehend guidance and genuinely teach users step by step. Nevertheless, as the context grows, even the ultra-large language models like GPT-4 and Claude-2 become constrained by the context length and gradually diverge. When approaching the limit, they exhibit a collapse phenomenon, revealing the final answer directly. By integrating a strong prior module with an output filter, our model avoids this issue. With merely 34B parameters, it achieves genuine tutoring ability in the truest sense.
\begin{table}[]
\caption{Evaluate The Tutoring Ability Of Different Model. We Will Place A Checkmark (\textcolor{red}{\CheckmarkBold}) Below The Corresponding Abilities That The Model Possesses. A Cross (\textcolor{green}{\XSolidBold}) Below Those not have.}\label{table:tab_tutor}
\centering
\setlength{\tabcolsep}{1mm}{
\begin{tabular}{lccc}
\hline
\textbf{Model} & \textbf{Tutor Style} & \textbf{Guided Output} & \textbf{w/o Answer} \\ \hline
StarCoder-15B\cite{li2023starcoder}  & \textcolor{green}{\XSolidBold} & \textcolor{green}{\XSolidBold} & \textcolor{green}{\XSolidBold} \\
CodeLlama-34B\cite{roziere2023code} & \textcolor{green}{\XSolidBold} & \textcolor{green}{\XSolidBold} & \textcolor{green}{\XSolidBold} \\
PanGu-Coder2-15B\cite{christopoulou2022pangu} & \textcolor{green}{\XSolidBold} & \textcolor{green}{\XSolidBold} & \textcolor{green}{\XSolidBold} \\
CodeFuse-CodeLlama-34B\cite{liu2023mftcoder} & \textcolor{green}{\XSolidBold} & \textcolor{green}{\XSolidBold} & \textcolor{green}{\XSolidBold} \\
\hline
C-Alpaca-13B    & \textcolor{green}{\XSolidBold} & \textcolor{green}{\XSolidBold} & \textcolor{green}{\XSolidBold} \\
Baichuan2-13B\cite{yang2023baichuan}     & \textcolor{red}{\CheckmarkBold} & \textcolor{green}{\XSolidBold} & \textcolor{green}{\XSolidBold} \\
ChatGLM3-6B\cite{zeng2022glm}     & \textcolor{red}{\CheckmarkBold} & \textcolor{green}{\XSolidBold} & \textcolor{green}{\XSolidBold} \\
Vicuna-13B\cite{chiang2023vicuna}      & \textcolor{red}{\CheckmarkBold} & \textcolor{green}{\XSolidBold} & \textcolor{green}{\XSolidBold} \\
Educhat-13B\cite{dan2023educhat}      & \textcolor{red}{\CheckmarkBold} & \textcolor{green}{\XSolidBold} & \textcolor{green}{\XSolidBold} \\
IBL-Tutoring-7B & \textcolor{red}{\CheckmarkBold} & \textcolor{green}{\XSolidBold} & \textcolor{green}{\XSolidBold} \\
\hline
SparkDesk-3.0& \textcolor{red}{\CheckmarkBold} & \textcolor{green}{\XSolidBold} & \textcolor{green}{\XSolidBold}\\
ERNIE-4.0 Bot & \textcolor{red}{\CheckmarkBold} & \textcolor{green}{\XSolidBold} & \textcolor{green}{\XSolidBold} \\ 
LaMDA\cite{thoppilan2022lamda} & \textcolor{red}{\CheckmarkBold} & \textcolor{green}{\XSolidBold} & \textcolor{green}{\XSolidBold} \\ 
Gopher & \textcolor{red}{\CheckmarkBold} & \textcolor{green}{\XSolidBold} & \textcolor{green}{\XSolidBold} \\ 
OPT-IML & \textcolor{red}{\CheckmarkBold} & \textcolor{green}{\XSolidBold} & \textcolor{green}{\XSolidBold} \\
Claude-2& \textcolor{red}{\CheckmarkBold} & \textcolor{red}{\CheckmarkBold} & \textcolor{green}{\XSolidBold} \\ 
GPT-3.5 Turbo   & \textcolor{red}{\CheckmarkBold} & \textcolor{green}{\XSolidBold} & \textcolor{green}{\XSolidBold} \\ 
GPT-4   & \textcolor{red}{\CheckmarkBold} & \textcolor{red}{\CheckmarkBold} & \textcolor{green}{\XSolidBold} \\ 
\rowcolor[HTML]{E7E6E6} 
ours-34B        & \textcolor{red}{\CheckmarkBold} & \textcolor{red}{\CheckmarkBold} & \textcolor{red}{\CheckmarkBold}
\\ \hline
\end{tabular}}
\end{table}


\subsection{Comparison Of Different Model Architectures}
In this section, we provide a visual comparison of different model outputs for the same question and system prompt, highlighting our strong cognitive abilities in the tutor role.
\subsection{Ablation Test}
We conducted a series of ablation experiments to investigate the significance and impact of several modules after data processing. We primarily simplified the evaluation of code capability, dialogue capability, and guidance capability. As mentioned earlier, we used the full 34B parameter model for code capability and guidance capability evaluations, and the 13B quantized model for dialogue capability evaluation.
\begin{table}[t!]
\caption{Ablation Test w/o Some Specific Parts On HumanEval(pass@1,code) And C-Eval(5-shot,chat).}\label{table:tab_ablation}
\centering
\setlength{\tabcolsep}{1mm}{
\begin{tabular}{lccc}
\hline
\textbf{Model}& \textbf{HumanEval}& \textbf{C-Eval}& \textbf{Tutoring Ability}\\
\hline
ours-unbroken & 75.10\%    & 50.60    & \textcolor{red}{\CheckmarkBold} 100\%          \\
ours-w/o Cutting \& Regularized & 76.20\%  & 51.10  & \textcolor{green}{\XSolidBold} 70\%            \\
ours-w/o Prior Module & 73.20\%    & 45.60  & \textcolor{green}{\XSolidBold} 80\%              \\
ours-w/o Filter      & 75.10\%   & 50.60 & \textcolor{green}{\XSolidBold} 98\% \\
ours-w/o 1st-stage & 33.60\%    & 38.80  & \textcolor{green}{\XSolidBold} 80\%              \\
ours-w/o 2nd-stage & 73.20\%  & 46.80 & \textcolor{green}{\XSolidBold} 75\%              \\
ours-w/o 3rd-stage & 58.20\% & 45.60 & \textcolor{green}{\XSolidBold} 70\%     
\\ \hline
\end{tabular}}
\end{table}

\ref{table:tab_ablation} reports the ablation result. In the evaluation results, we found that the first phase of fine-tuning has the most significant impact on the model's dialogue and code capabilities, while the third phase of fine-tuning and the constraint term have the most significant influence on the model's guidance-oriented output. This is mainly because the first phase contains a large number of book dialogue and code data, and a clean dataset can significantly improve the model's general capabilities. The third phase greatly enhances the model's step-by-step output capabilities and context coherence through multi-turn dialogue data, while the regularization constraint effectively restricts guidance-oriented output further.
\section{Conclusions}
In this paper, we illustrate the challenges encountered by LLMs in the field of education. The limitations of both domain expertise and the understanding of the tutor hinder the effective implementation of a tutor using LLMs. To address this, we propose a three-step fine-tuning method via a carefully designed priori module. The entire end-to-end system design encompasses an overlap neural network for data processing at the input end, filter for temporal text at the output end, and robust feature constraints throughout. Our method exhibits strong robustness and high accuracy, achieving state-of-the-art performance in code evaluation, while possessing excellent tutor teaching capabilities. The system can digest and break down knowledge from textbooks and provide step-by-step incremental guided output to students, making it easier for them to comprehend. Furthermore, our method demonstrates notable transferability, as the ideas of data preprocessing, step-wise fine-tuning, and feature constraints can be applied to a broader range of vertical domains.

\bibliography{references}
\bibliographystyle{IEEEtran}

\end{document}